\documentclass[10pt]{article}
\usepackage[a4paper, total={6.5in, 10in}]{geometry}
\usepackage{wrapfig}
\usepackage{times}
\usepackage{epsfig}
\usepackage{graphicx}
\usepackage{amsmath}
\usepackage{amssymb}
\usepackage{verbatim}
\usepackage{mwe}
\usepackage{subfig}
\usepackage[font=small, labelfont=bf]{caption} 
\usepackage{cuted}
\usepackage{capt-of}
\usepackage{dblfloatfix}
\usepackage{graphicx}
\usepackage{amsmath}
\usepackage{amssymb}
\usepackage{xcolor}
\usepackage{soul}
\usepackage{booktabs}
\usepackage{multirow}
\usepackage{multicol}
\usepackage{caption}
\usepackage{tabularx}
\usepackage{float}

\usepackage[breaklinks=true,bookmarks=false]{hyperref} 

\date{}
\begin{document}

\title{Vision in adverse weather: Augmentation using CycleGANs with various object detectors for robust perception in autonomous racing}


\author{Izzeddin Teeti\thanks{Autonomous Driving $\&$ Intelligent Transport Group, Oxford Brookes University} 
\and Valentina Mușat\thanks{Oxford Robotics Institute, University of Oxford} 
\and Salman Khan\thanks{Visual Artificial Intelligence Laboratory, Oxford Brookes University}
\and Alexander Rast\footnotemark[1]
\and Fabio Cuzzolin\footnotemark[3]
\and Andrew Bradley\footnotemark[1] \footnote{for email correspondence: abradley@brookes.ac.uk}
}

\maketitle



\begin{abstract}
   \vspace{-1mm}
In an autonomous driving system, perception - identification of features and objects from the environment - is crucial. In autonomous racing, high speeds and small margins demand rapid and accurate detection systems. During the race, the weather can change abruptly, causing significant degradation in perception, resulting in ineffective manoeuvres. In order to improve detection in adverse weather, deep-learning-based models typically require extensive datasets captured in such conditions - the collection of which is a tedious, laborious, and costly process. However, recent developments in CycleGAN architectures allow the synthesis of highly realistic scenes in multiple weather conditions. To this end, we introduce an approach of using synthesised adverse condition datasets in autonomous racing (generated using CycleGAN) to improve the performance of four out of five state-of-the-art detectors by an average of 42.7 and 4.4 mAP percentage points in the presence of night-time conditions and droplets, respectively. Furthermore, we present a comparative analysis of five object detectors - identifying the optimal pairing of detector and training data for use during autonomous racing in challenging conditions.

\end{abstract}

\section{Introduction}

Autonomous driving is a current focus of many governments, startups, already-established corporations, and research labs \cite{CBIInsights}. Many studies aim to create an SAE level 4 or level 5 autonomous vehicle (AV) \cite{SAE2018}, able to operate in unstructured environments and adverse conditions. From the AV's perspective, an adverse condition is \textit{any} condition with which the detection model has never experienced during training. Of particular interest are the effects of weather and lighting effects that change the appearance of objects, such as droplets (which can occlude objects and blur images), or night driving.

Research in AVs has stimulated the development of autonomous racing, where cars designed for racing compete fiercely on racetracks, potentially under varied weather conditions which may be subject to rapid changes. The close, competitive nature of racing amplifies the effects of any performance gains (or weaknesses) in the AV control system in a safe, controlled test environment \cite{Culley2020}. Autonomous racing is therefore an ideal application to test the extreme operating limits of AVs, accelerating the development process of AV algorithms - including object detection in adverse conditions.

Two main approaches exist to AV control systems. The first is an `end-to-end' AI-based approach that takes visual inputs and outputs control commands \cite{Bansal2018}. The end-to-end nature of such systems, however, makes them black boxes: they lack explainability and are difficult to debug. Furthermore, they require an extensive amount of labelled training data; however, gathering such data is time-consuming and expensive. The second approach breaks down the AV control system into a pipeline of interacting subsystems including perception, decision making, and control \cite{Badue2021} - enhancing explainability, debugging, and a higher confidence in the ability of the autonomous vehicle to handle a variety of road scenarios. Both approaches, however, are vulnerable to adverse conditions, as the latter distorts the visual input upon which they depend.

AVs see (perceive) the world through visual inputs, acquired by sensors such as (optical) cameras and LiDAR - both of which are susceptible to adverse conditions including snow, rain, fog, and night \cite{Mohammed2020ThePS}. Rain droplets can accumulate on camera lenses, occluding objects and agents, or distorting their shapes. If a LiDAR beam intersects a rain droplet at a short distance from the transmitter, the raindrop can reflect enough of the beam back to the receiver to cause the raindrop to be detected as an object \cite{Espineira2021}. Night driving is particularly challenging for optical sensors; the level of reflected light from target objects or agents to the camera sensor is low, causing dim or barely recognisable objects. 

Perception tasks, including object detection, semantic segmentation, action recognition, and trajectory prediction are hampered by poor visual inputs caused by adverse weather, resulting in missed and incorrect detections \cite{Mohammed2020ThePS} - of critical importance for AVs to safely 'see' other road users. Errors in detections will propagate through downstream tasks including path planning and control, causing inaccurate and unreliable manoeuvres and compromising the safety of other road users. In the particular application of AV racing, a small error in perception may result in a considerable deviation from the optimal path \cite{KATRAKAZAS2015416} and a loss of time or even penalties or accidents. In both on-road and racing applications, the AV industry has a critical and urgent need to mitigate or deal with the significant degradation of perception caused by adverse weather conditions \cite{AirSim, Fursa2021}.

Since both end-to-end and independent-subsystems approaches typically use machine learning models to understand the visual inputs, one way to improve them in adverse weather is to retrain those models on more data that contains adverse weather conditions. However, obtaining this data is time-consuming, expensive, and at the mercy of the weather. It also introduces a very costly and laborious process of (manually) labelling the collected data for different tasks like detection or segmentation. To overcome this challenge, researchers have suggested synthesising adverse weather datasets using simulators \cite{Zadok2019} or simple augmentation algorithms \cite{Fursa2021}. However, the applicability of these approaches for enhancement of object detection in a real-world setting is questionable, due to the lack of visual realism in the imagery.

A recent promising approach is harnessing CycleGAN to generate more realistic weather augmentations using style transfer. A small number of studies have explored the use of CycleGAN to improve perception performance in adverse weather in the context of on-road autonomous driving tasks \cite{Porav2020, Uricar2019, Musat_2021_ICCV}, but as yet, no studies have attempted to exploit this potential in the context of autonomous racing.

Furthermore, various object detection models have been proposed for real-time use in AV applications \cite{jocher2021YOLOv3, jocher2020YOLOv5, ren2015faster, tan2020efficientdet}, however few studies have compared different detectors for use in autonomous driving applications \cite{Gupta2021} - let alone with a detailed comparison of the detection performance and latencies introduced.


This study represents the first attempt to harness CycleGAN-based weather appearance transfer to generate realistic training data for multiple adverse weather conditions in autonomous racing, and explores the performance of five leading object detectors for real-time object detection in adverse conditions. Thus, the contributions of this study are:

\begin{enumerate}
    \setlength\itemsep{0.0em}
    \vspace{-0.2cm}
    
    \item An approach to using individual \textit{synthesised} adverse weather conditions to improve detection performance under multiple \textit{real} weather conditions in autonomous racing. \vspace{-0.1cm} 
    \item Performance improvements for four state-of-the-art object detectors under multiple real adverse weather conditions by training them on synthesised adverse weather datasets. \vspace{-0.1cm}
    \item Benchmark results for five different object detectors, a comparative analysis of their performance in terms of accuracy and speed in both sunny and adverse conditions, and the identification of the optimal detector for the specific task of object detection for autonomous racing cars. \vspace{-0.1cm}

\end{enumerate}

\section{Related work}

\subsection{Object detection in Autonomous Racing}

Traffic cones are used to regulate traffic and to mark racing circuit layout, therefore detecting them is crucial for autonomous driving and autonomous racing. The earliest cones detection methods proposed simple colour-based techniques using filters and thresholds to find cones \cite{Jin2012RecognitionOT}. Generally, these techniques are quite unreliable and are characterised by many false positives \cite{Hashemi2016TemplateMA}, because they ignore a fundamentally distinctive feature of cones: their shape. Template matching methods, which account for both colour and shape of cones, were also attempted for cone detection \cite{Yong2015RealtimeTC, Zeilinger2017}. Although they show accuracy improvements, their results are poor when cones are slightly deformed, partially occluded, or appear at multiple scales. Feature-based detectors have also been explored by \cite{TIAN2018}, which used Histogram of Oriented Gradients (HOG) \cite{Dalal2005} to detect cones. 

In the current decade, Convolutional Neural Networks (CNNs), have proven to be highly accurate for many perception tasks, such as recognition, localisation, segmentation, and, most importantly, detection \cite{Khan2020}. CNNs are characterised by multiple feature extraction stages that can automatically learn representations from the data, thus, trained on the right dataset, CNN-based models can detect objects even when occluded, deformed, or rescaled.

For cone detection in a racing context, papers such as \cite{Qie2020ConeDA} employed a colour-based detector, whereas \cite{TIAN2018} used an HOG representation. Many systems, however, including \cite{Dhall2019, Strobel2020AccurateLV, Culley2020}, use different versions of the well-known YOLO detector \cite{Redmon2016YouOL} because of its speed, accuracy, and reliability. A variety of object detectors were used in AV racing, however, there is no clear study that makes comparative analysis between these models - and therefore an investigation into the ideal detector, considering detection accuracy and speed, would be a useful contribution to the field. 

Despite being the most accurate and most commonly used class of detectors, however, CNN-based models display a significantly degraded performance in adverse weather conditions \cite{Fursa2021}. While this can be mitigated by training the model on datasets that contain adverse weather, the collecting and labelling of such datasets is expensive, laborious, and at the mercy of the prevailing weather conditions. An alternative solution is generating such data using simulators, physics models, or style transfer techniques like CycleGAN. However, augmentation affects the performance of different detectors in different ways. This paper makes a comparative analysis between different object detectors in terms of accuracy, speed, and their behaviour against augmentation.  

\subsection{Synthetic weather generation}

Numerous approaches were proposed to generate synthetic adverse weather images. Simple approaches have utilised open-source frameworks like OpenGL \cite{Praveen2017}, and image augmentation libraries including \cite{Fursa2021}
, however, they lack visual realism. More sophisticated methods were proposed to generate more realistic adverse weather images. They harness physics models to model the dynamics of raindrops \cite{Creus2013}, as well as, the behaviour of the light going through them \cite{Bernard2013}. Despite their improved realism, such models demand in-depth knowledge of many controlling parameters, which, in turn, are empirically determined.

Due to the difficulty in generating synthetic data using simple augmentations and physics engines (low realism, high number of free parameters), the autonomous driving research has seen an increase in the use of GANs and its variants to address the issue of data scarcity in adverse weather conditions. As collecting data in harsh conditions and providing ground truth for it is an expensive and time-consuming task, these approaches have been employed to generate datasets in various conditions based on existing datasets and their corresponding ground truth. The first to explore this avenue for autonomous driving was the work of \cite{porav2018}, which generates new images with snow and various light conditions for the purpose of improving feature matching. Other applications focus on improving vehicle detection during nighttime \cite{Lin2020}, object detection and semantic segmentation \cite{Ostankovich2020}.

Other articles have used CycleGAN in the context of autonomous driving tasks \cite{Porav2020, Uricar2019, Musat_2021_ICCV}. The latter is the closest to our work; using CycleGAN to create fake adverse conditions for road-going AVs in general rather than racing AVs. While the augmentations demonstrated an improvement in perception performance in adverse weather, only one detector was used, and no consideration was given to how different detectors may perform when augmented training data is used, or their suitability for real-time use on-board a vehicle.

\subsection{Adverse weather in autonomous racing}


Adverse weather affects both the vehicle dynamics (due to, e.g., ice, water puddles or soil on the road surface) and the perception stack of the autonomous vehicle (due to both detection occlusions and changes in the road surface). Different methods have been proposed to address the problem of adverse weather from a perception point of view. \cite{Zadok2019} who improved an end-to-end model by tackling weather appearance with the use of the existing Airsim simulator \cite{AirSim} to generate various time-of-day appearances. The method is scalable, however, it generates data only in day time, and the generated images are not realistic, thus the data may still be susceptible to differences between simulation and real world. Although such models might perform well on simulation images, they may fail on images coming from real-world unconstrained environments. \cite{Fursa2021} who employed existing simple open-source augmentation libraries to generate adverse weather datasets, but likewise potentially suffers from similar issues. Other methods including \cite{Porav2020, Desoiling_Dataset} used GANs to improve tasks downstream from perception in urban and rural autonomous driving. To the best of our knowledge, this study represents the first attempt to harness CycleGAN-based weather appearance transfer to generate realistic datasets for multiple adverse weather conditions, in order to improve object detection in real adverse conditions in the context of autonomous racing.



\section{Data}

\subsection{Real-world dataset}\label{sec:real_dataset}

\begin{figure}[t]
 \centering
 \subfloat[Real sunny weather]{\includegraphics[width=0.17\textwidth]{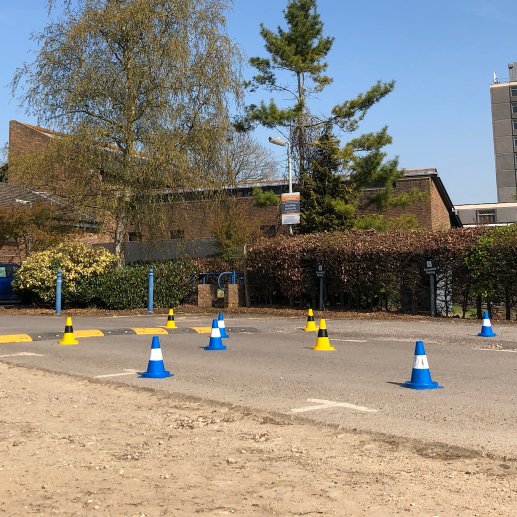}\label{fig:real_sunny}}%
 \hspace{0.4mm}
 \subfloat[Real night]{\includegraphics[width=0.295\textwidth]{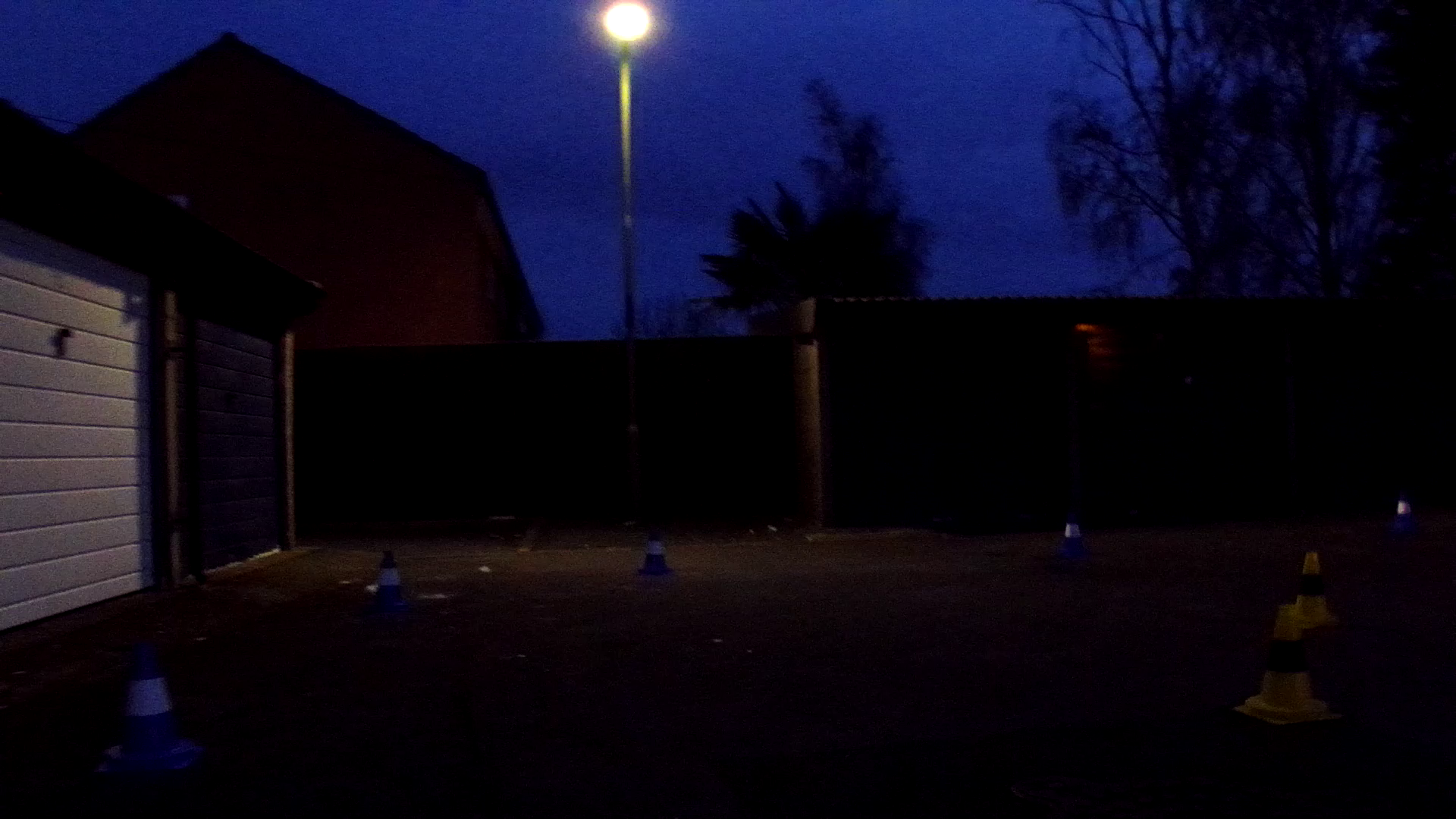}\label{fig:real_night}}
 \hspace{0.4mm} 
 \subfloat[Real droplet]{\includegraphics[width=0.30\textwidth]{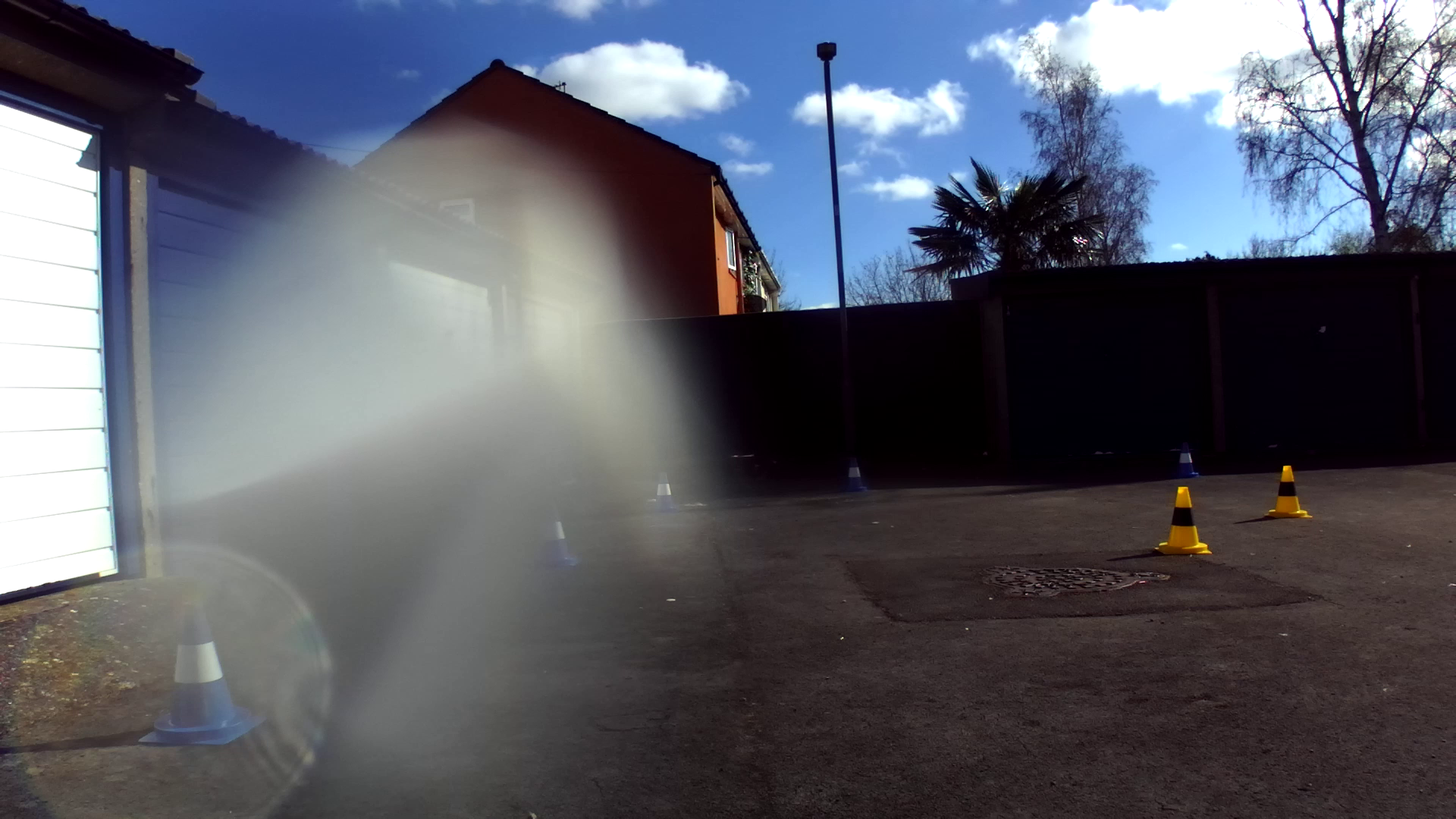}\label{fig:real_rain}}%
 \vspace{-2mm}
 \caption{Real datasets}%
 \label{fig:real_dataset}%
\end{figure}

For the object detection task, we used three real datasets of various conditions (Figure \ref{fig:real_dataset}) that were captured and annotated from an autonomous racing vehicle's vision sensor. The first dataset consists of more than $10,125$ cones captured in sunny conditions; the second ($1,548$ cones) is captured under low light conditions after sunset; the final dataset ($1,197$ cones) was captured in conditions where adherent droplets on lens completely occlude and distort some of the cones (Figure \ref{fig:real_rain}, in contrast to the figures shown in \cite{Dhall2019}, which show no drastic changes in the cones' appearance). 

We used the first dataset to train and test the five object detectors ($70\%$ training, $30\%$ testing), and kept the last two datasets to test the detectors against out-of-sample adverse illumination and rainy conditions. We exclusively focus on two object classes: blue and yellow cones.

\subsection{Synthetic dataset}

\begin{figure}[tp]
\centering
\includegraphics[width=0.45\textwidth]{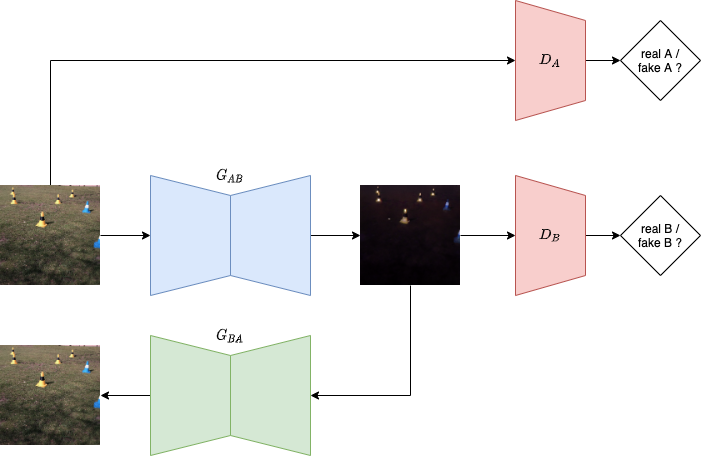}
\caption{CycleGAN architecture}
\label{fig:cycleGAN_arch}
\end{figure}

Since obtaining data with ground truth in adverse conditions is a time- and effort-intensive task, we employ a CycleGAN architecture \cite{Zhu2017} (Figure \ref{fig:cycleGAN_arch}) to generate synthetic night and rainy (droplet) conditions with the datasets that we already have. A CycleGAN is an image-to-image translation model that learns to map images from domain $A$ to domain $B$ while preserving the structure of the image. It is a variant of the original GAN model, one in which two generators are trained simultaneously (one to learn the mapping from $A\rightarrow B$ and the other the mapping from $B\rightarrow A$), while enforcing a cycle-consistency \cite{Zhu2017}. 

Cycle-consistency is enforced by training each GAN on the output of the other: the second generator receives as input the output of the first generator (image in domain $B$)  and is trained to reconstruct the original image (image in domain $A$), by applying a reconstruction loss between the real image in domain $A$ and its reconstruction $G_{BA}(G_{AB}(A)) = G_{BA}(\hat{B}) = \Tilde{A} \approx A$. Similarly, the first generator receives as input the output of the second generator (image in domain $A$), and is trained to reconstruct the original image (image in domain $B$), by applying a reconstruction loss between the real image in domain $B$ and its reconstruction $G_{AB}(G_{BA}(B)) = G_{AB}(\hat{A}) = \Tilde{B} \approx B$.

In order to generate synthesised images that contain night or droplets conditions, we trained off-the-shelf CycleGAN implementation \cite{Zhu2017} on two (non-racing) datasets from a mixture of resources. The first CycleGAN was trained on frames from a YouTube video 
captured at night, in London's streets \cite{youtube_night} (Figure \ref{fig:train_night}), while the second CycleGAN was trained on in-house acquired images that contain droplets (Figure \ref{fig:train_droplet}), both were trained at a resolution of $512\times512$. Then, we used them to transfer the night or droplet styles to the sunny images, ending up with 911 `fake' night images (Figure \ref{fig:fake_night}) and 800 `fake' droplet images (Figure \ref{fig:fake_droplet}), both having the same ground truth labels as the original sunny images. Since these fake conditions are meant for training the detectors, we split them into $70\%$ training set (637 fake night and 560 fake droplet cone images) and $30\%$ validation set (274 fake night and 240 fake droplet cone images). We used the training set to train the detectors and the validation set to measure their performance while training. Training the detectors with fake adverse conditions then testing them on real adverse conditions allows us to verify that fake generated images are a good proxy for real adverse conditions images.

\subsection{Qualitative analysis of the generated datasets} \label{sec:fid}

FID measure is the 2-Wasserstein distance \cite{OLKIN1982} between two multidimensional Gaussian distributions in the feature space:  ${\displaystyle {\mathcal {N}}(\mu _{s},\Sigma _{s})}$ the distribution of some neural network features of the synthesized images generated by the GAN and ${\displaystyle {\mathcal {N}}(\mu _{r},\Sigma _{r})}$ the distribution of the same neural network features from the real images used to train the GAN. Mathematically, ${\text{FID}}=||\mu_{r} -\mu_{s}||_{2}^{2}+\operatorname {trace} (\Sigma_{r} +\Sigma_{s}-2(\Sigma_{r} \Sigma_{s})^{1/2})$ \cite{OLKIN1982}.

FID measure is $83.56$ and $138.39$ for night and droplet conditions, respectively. The lower the value of FID measure, the closer the distribution of the generated dataset to the that for the real dataset, thus, the more similar is the generated dataset to the real dataset. Therefore, the results suggest that our artificially generated night is more similar to a real night than our generated droplet images are to real images with droplets. 

\section{Object Detectors}

We decided to test our concept on five state-of-the-art object detection models, not only to validate our concept, but also to run a comparative analysis between these detectors and identify the optimal one, which is crucial information for autonomous racing research. The following five object detectors were chosen due to their high speed, high accuracy, or because they can be efficiently deployed for autonomous driving applications,


\textbf{YOLOv3-tiny} has fewer convolution layers than YOLOv3 \cite{redmon2018YOLOv3}. With fewer layers, the tiny version is claimed to be the fastest detection model \cite{YI201917}, making it ideal for high-speed applications like autonomous racing. Counterintuitively, this high speed does not significantly compromise accuracy for our problem which only contemplates two classes (blue and yellow cones). 

\textbf{Scaled YOLOv4} \cite{Wang_2021_CVPR} was chosen because it scales the resolution, width, and depth (number of layers) of the original YOLOv4 \cite{bochkovskiy2020YOLOv4}, making it more efficient and robust. Furthermore, their authors claim it delivers the highest accuracy on the Microsoft COCO (Common Objects In Context) dataset \cite{lin2015microsoft}. A modified, effective and real-time version of YOLOv4 was implemented in \cite{Cai2021} for object detection in autonomous driving.

\textbf{YOLOv5s} \cite{jocher2020YOLOv5} is the small version of YOLOv5. It differs from YOLOv4 by being computationally more efficient in terms of storage space and training time. In terms of accuracy and inference speed, however, there is no quantitative evidence on which is better. Consequently, YOLOv5s was chosen to be compared with other detectors in our autonomous racing domain. \cite{gu2021realtime} adopted YOLOv5 medium for real-time detection in autonomous driving, their system achieved a balance between accuracy and latency as it scored the second place at the Streaming Perception Challenge (Workshop on Autonomous Driving at CVPR 2021), while running in real-time. Additionally, \cite{benjumea2021yoloz} implemented a modified version of YOLOv5 called YOLO-Z to detect small objects in the context of autonomous racing.

\textbf{Faster-RCNN} \cite{ren2015faster} is a well-known Region Proposal Network (RPN)-based object detection paradigm. The model was initially proposed with a VGG \cite{simonyan2014very} backbone; here we used a Feature Pyramid Network (FPN) \cite{fpn2016} backbone because of its superior performance \cite{wu2019detectron2}. Faster-RCNN was extended to stereo inputs in \cite{Li_2019_CVPR} to build a 3D object detector for autonomous cars that achieved state-of-the-art results.


\textbf{EfficientDet} \cite{tan2020efficientdet} applies weighted bi-directional FPN and compound scaling to the EfficientNet classifier \cite{tan2019efficientnet}, resulting in a highly computationally efficient detector. Given the principal goal of detecting cones in real-time, we selected EfficientDet-D0 which is the lightest and most efficient among the 9 EfficientDet models.

\section{Experiments}

\subsection{Experimental Method} \label{sec:method}

The general approach of the experiments is to train the five detectors on sunny weather to then test them on real adverse conditions. After that, we retrain them on fake adverse conditions then test them again on real adverse conditions. The goal is to record the difference in performance between models trained on (real) sunny condition data versus models trained on (fake) adverse conditions. If the latter proved to be better, then we would have demonstrated that our (synthesised) datasets are an effective proxy for (real) adverse condition data in the domain of autonomous racing. Furthermore, experiments aim to analysis the response of detectors to augmentation and identify the detector with best response. Additionally, the inference time of detectors is recorded to identify their suitability to real-time operation. 

\begin{figure}[h]
 \centering
 \subfloat[Real sunny (original image)]{\includegraphics[width=0.23\textwidth]{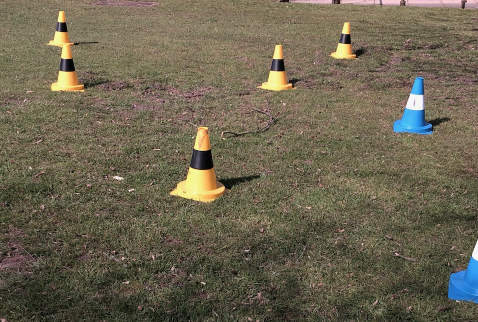}\label{fig:real_sunny2}}\\
 \vspace{-3mm}
 \subfloat[Sample used to train CycleGAN to generate night]{\includegraphics[width=0.23\textwidth]{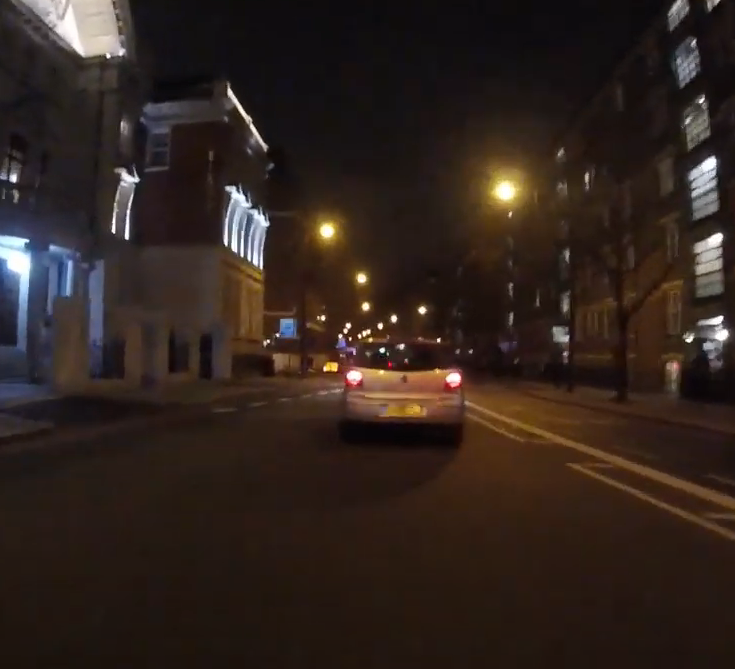}\label{fig:train_night}}
 \hspace{0.5mm}
 \subfloat[Sample used to train CycleGAN to generate droplets]{\includegraphics[width=0.23\textwidth]{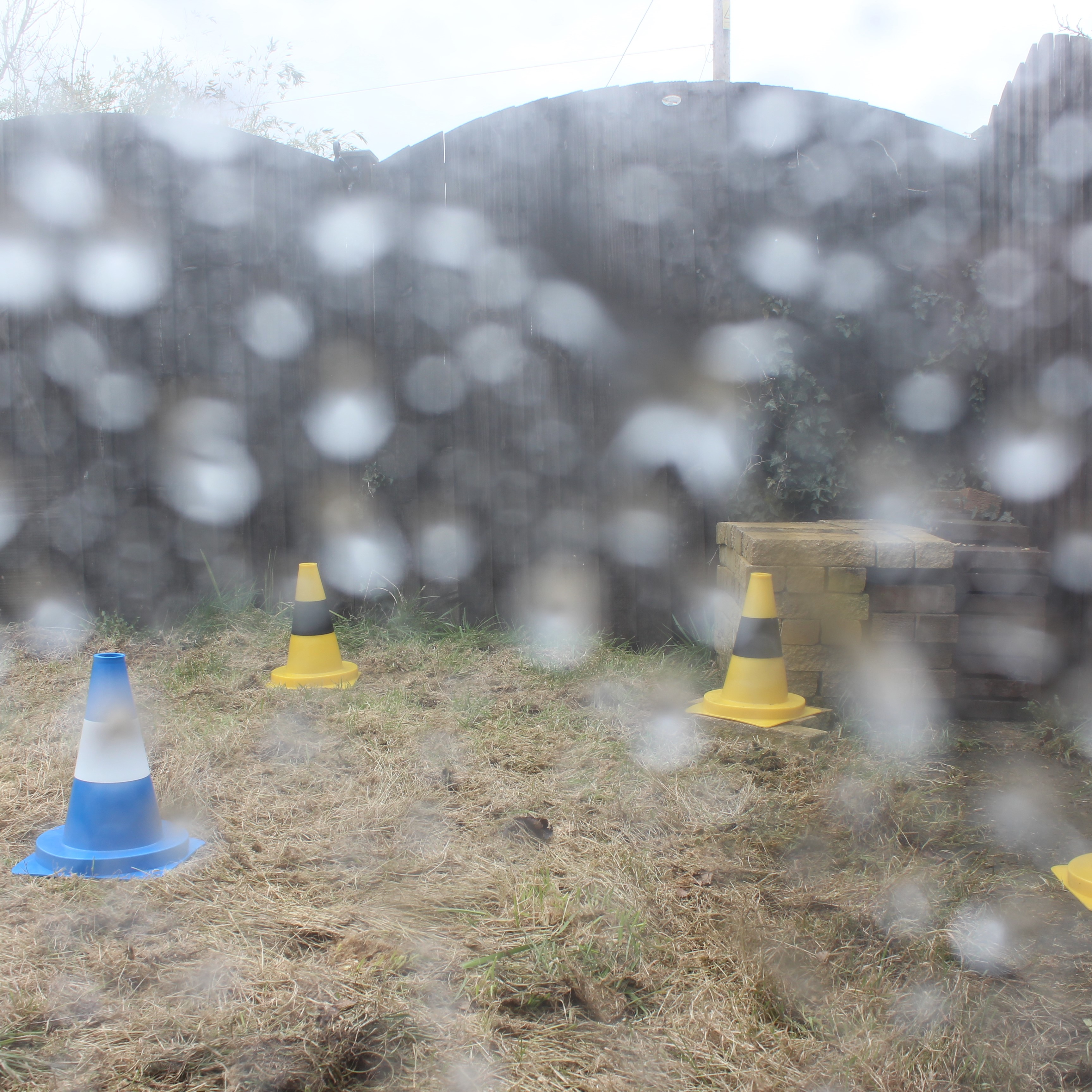}\label{fig:train_droplet}}\\
 \vspace{-3mm}
 \subfloat[Generated night]{\includegraphics[width=0.23\textwidth]{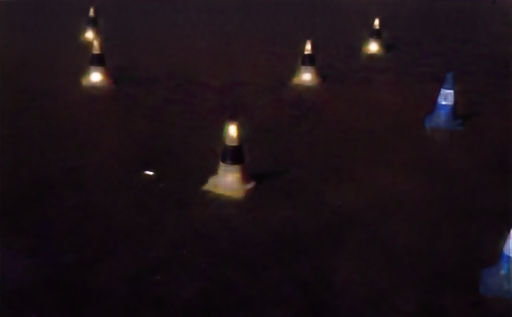}\label{fig:fake_night}}
 \hspace{0.5mm}
 \subfloat[Generated droplets]{\includegraphics[width=0.23\textwidth]{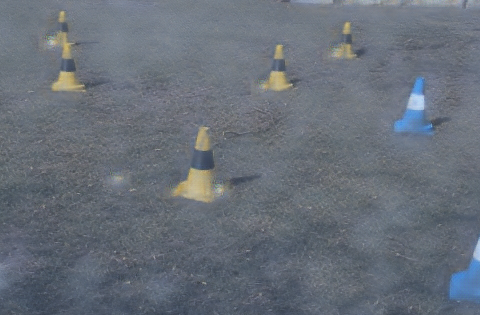}\label{fig:fake_droplet}}
 \vspace{-2mm}
 \caption{\textbf{a)} An original image in sunny weather, which is the input of the CycleGAN. \textbf{b, c)} Examples of images used to train the CycleGAN to generate night and droplet conditions. \textbf{d, e)} the generated images based on the input image (a) with the style of the training images (b) and (c).}%
 \label{fig:generated_conditions}%
\end{figure}

Our experiments were divided into four stages. Firstly, all five detectors were trained on \textbf{sunny weather} data (see Table \ref{tab:results}, first row). This set the baseline against which the rest of the models were compared. Secondly, the detectors were trained on \textbf{Sunny and Fake Night} (see Table \ref{tab:results}, second row, and Figure \ref{fig:fake_night}). Thirdly, they were trained on \textbf{Sunny and Fake Droplet} data (third row and Figure \ref{fig:fake_droplet}). Finally, they were trained on \textbf{Sunny, Fake Night and Fake Droplet} data (forth row). In each stage the trained detectors were tested on \textbf{Sunny}, \textbf{Real Night}, and \textbf{Real Droplet} data (see Section \ref{sec:real_dataset}).

In total, \textbf{5} detectors were trained on \textbf{4} sets of data and tested on \textbf{3} sets of data, yielding a total of \textbf{60} tests ($5\times4\times3$). The results are reported in Table \ref{tab:results}, while a performance comparison across stages is shown in Figure \ref{fig:imp1}.

\subsection{Experimental Setup} \label{sec:exp_set}

In all our experiments the training was conducted on a GPU machine equipped with 4 Nvidia GTX 1080 GPUs with 12GB VRAM each. To perform fair comparisons between all five object detectors we fixed the value of several hyperparameters, including number of epochs (to 300), batch size (to 64), and input image size (to $512\times512$ pixels). 
For testing, we used an in-car PC mounted on a racing car equipped with a single Nvidia GTX 1050 TI GPU, with a batch size of 1 and an image size of $512\times512$ pixels. 

Performance was measured using the standard mean Average Precision (mAP), with a constant Intersection over Union (IoU) threshold of 0.5 (Table \ref{tab:results}). In addition, we measured the computational efficiency by recording for each detector the floating point operations per second (FLOPS) and the inference time in frames per seconds (fps) (last two rows of Table \ref{tab:results}).

\begin{table*}[h]
\begin{center}
\centering
  {
\begin{tabular}{cl|ccccc}
\toprule
\multicolumn{2}{c|}{\textbf{Data splits}}                                                   & \multicolumn{5}{c}{\textbf{Detection model mAP}}                                                 \\ 
\textbf{Training set}                              & \multicolumn{1}{c|}{\textbf{Test set}} & \textbf{YOLOv3} & \textbf{SYOLOv4} & \textbf{YOLOv5} & \textbf{FRCNN} & \textbf{EfficientDet} \\ \hline
\multirow{3}{*}{Sunny}                             & Sunny                                  & 0.89            & 0.93             & 0.95            & \textbf{0.97}  & 0.45                  \\ \cline{2-7}
                                                   & Real Night                             & \textbf{0.43}   & 0.21             & \hypertarget{first}{0.22}            & 0.23           & 0.24                  \\ \cline{2-7}
                                                   & Real Droplet                           & 0.53            & 0.54             & 0.53            & \textbf{0.57}  & 0.43                  \\ \hline
\multirow{3}{*}{Sunny + Fake Night}                & Sunny                                  & 0.89            & 0.89             & \textbf{0.97}   & 0.96           & 0.47                  \\ \cline{2-7}
                                                   & Real Night                             & 0.62            & 0.39             & \textbf{0.73}   & 0.67           & 0.55                  \\ \cline{2-7}
                                                   & Real Droplet                           & 0.57            & 0.55             & 0.58            & \textbf{0.60}  & 0.53                  \\ \hline
\multirow{3}{*}{Sunny + Fake Droplet}              & Sunny                                  & 0.89            & 0.88             & \textbf{0.96}   & 0.95           & 0.36                  \\ \cline{2-7}
                                                   & Real Night                             & \textbf{0.57}   & 0.33             & 0.25            & 0.30           & 0.26                  \\ \cline{2-7}
                                                   & Real Droplet                           & 0.54            & 0.51             & 0.55            & \textbf{0.62}  & 0.36                  \\ \hline
\multirow{3}{*}{Sunny + Fake Night + Fake Droplet} & Sunny                                  & 0.89            & 0.86             & \textbf{0.97}   & \textbf{0.97}  & 0.37                  \\ \cline{2-7}
                                                   & Real Night                             & 0.68            & 0.66             & \hypertarget{second}{\textbf{0.74}}   & 0.71           & 0.25                  \\ \cline{2-7}
                                                   & Real Droplet                           & 0.58            & 0.55             & \textbf{0.61}   & 0.60           & 0.37                  \\ \bottomrule
\multicolumn{2}{c|}{\textbf{Speed (fps) $\uparrow$}}                                        & \textbf{74.63}  & 8.77             & 13.68           & 3.01           & 26.53                 \\ \cline{1-7}
\multicolumn{2}{c|}{\textbf{FLOPS (B) $\downarrow$}}                                        & 12.9            & 109              & 16.4            & 180            & \textbf{2.5}          \\ \bottomrule
\end{tabular}
}
\end{center}
\vspace{-0.4cm}
\caption{Detailed comparative analysis of mAP for all five object detectors (tested on Sunny, Real Night, and Real Droplet) when trained on four different training sets. The aim of that is to show the difference of detectors' performance on real adverse conditions when trained on fake adverse condition against that when trained on sunny condition only. mAP is measured at a standard IoU threshold of 0.5. The plus (+) shows the concatenation of real and augmented data for training. The best model of each training-testing stage (each row). Frame rate (in fps) and FLOPS (in billions) comparison of all five object detectors, tested using an AV's on-board PC with GTX 1050 TI.}
\label{tab:results}
\end{table*}



\section{Results}

\subsection{Impact of synthetic adverse conditions} 

\begin{figure*}[h]
\centering
\includegraphics[width=\linewidth]{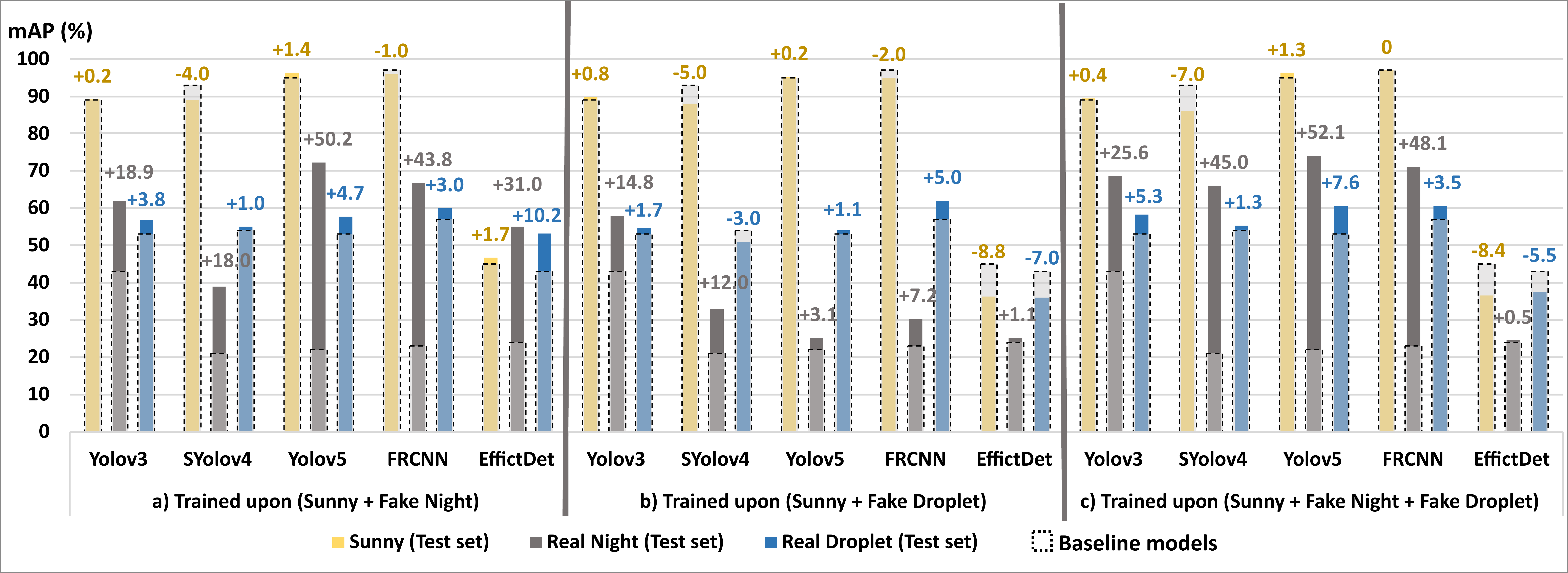}
\caption{Dashed bars show the baseline performance (mAP) of 5 models trained on \textbf{Sunny} and tested on Sunny, Real Night, and Real Droplet ($1^{st}$ row of Table \ref{tab:results}). The coloured continuous bars and coloured numbers show the improvement (+ve value) or degradation (-ve value) of performance caused by retraining the models on \textbf{a) Sunny+Fake Night. b) Sunny+Fake Droplet, c) Sunny+Fake Night+Fake Droplet}, ($2^{nd}$, $3^{rd}$, and $4^{th}$ rows of Table \ref{tab:results}).}
\label{fig:imp1}
\end{figure*}

Detector performance (mAP) over the 60 experiments discussed in Section \ref{sec:method} is reported in Table \ref{tab:results}. Figure \ref{fig:imp1} shows the difference in object detection performance of the models trained on synthetic adverse conditions dataset against that of their baseline versions (trained only on Sunny weather). Figure \ref{fig:imp1}a compares the models trained upon Fake Night to their corresponding baseline ($1^{st}$ and $2^{nd}$ rows in Table \ref{tab:results}). It is clear that, after training the detectors on Fake Night, their performance on Real Night improves by an average of 32 percentage points (pp). Perhaps surprisingly, training on Fake Night not only improved the detectors' performance on real night, but also on Real Droplet by an average of 4.5 pp. A possible reason for this is that by introducing Fake Night in the training dataset, the models were trained to infer more diverse images that are out-of-sample of the original Sunny Weather training data. 

Figure \ref{fig:imp1}b compares the models trained upon Fake Droplet to their respective baseline ($1^{st}$ and $3^{rd}$ rows in Table \ref{tab:results}). When trained on Fake Droplet, YOLOv3, YOLOV5 and Faster-RCNN show improvements on Real Droplet, and just like the previous stage, they also show improvements on the other real adverse condition (Real Night). Counterintuitively, training on Fake Droplets slightly \textit{reduced} performance in Real Droplet conditions and \textit{improved} detections in Real Night using YOLOv4 and EfficientDet.

\begin{figure}[h]
 \centering
 \subfloat[Before training on synthetic night and synthetic droplet (mAP=0.22)]{\includegraphics[width=0.49\textwidth]{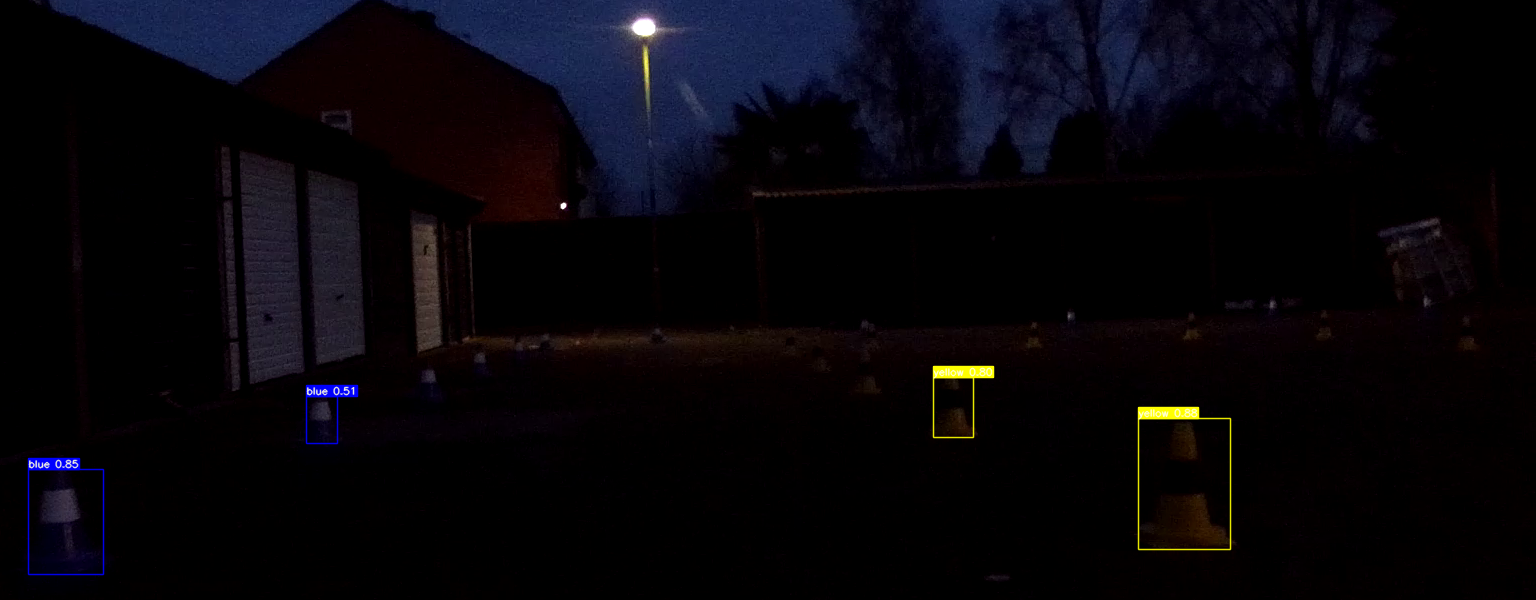}\label{fig:before_aug}} \vspace{-2mm}
 \subfloat[After training on synthetic night and  synthetic droplet (mAP=0.74)]{\includegraphics[width=0.49\textwidth]{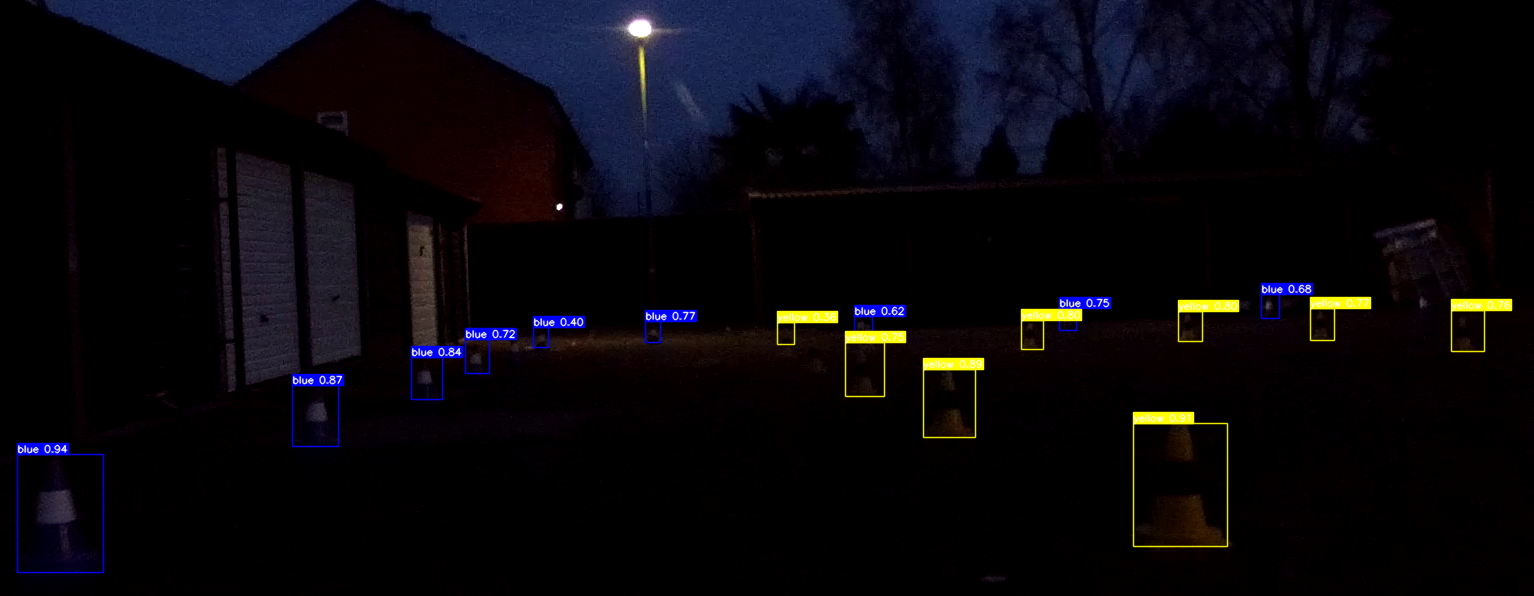}\label{fig:after_aug}}%

 \caption{Impact of augmentation on enhancing night detections using YOLOv5}%
 \label{fig:QualRes}%
\end{figure}

Finally, Figure \ref{fig:imp1}c shows the difference between The `monolithic' versions of the models (those trained on Sunny, Fake Night and Fake Droplet) and their related baseline ($1^{st}$ and $4^{th}$ rows in Table \ref{tab:results}). The `monolithic' versions exhibit slightly better performance on real adverse conditions than those trained on a single fake adverse condition. When tested on Real Night, the monolithic models (excluding EfficientDet) show a mean performance improvement of 42.7 pp, while the models trained only on Fake Night show a mean increase of 32.7 pp. In Real Droplet conditions, the monolithic models show a mean increase of 4.4 pp, while the models trained only on Fake Droplet show a slight degradation of 1.2 pp. Figure \ref{fig:QualRes} show the qualitative results of YOLOv5 night detections before and after being trained on synthetic night.

The average improvements on all conditions for night is higher than that for droplets. This is due to the baseline models having lower performance on Fake Night than on Fake Droplets, so there was more room for improvement in the former. Furthermore, this agrees with the fact that the synthetic Fake Night data is more visually similar to the Real Night data than the Fake Droplet data is to Real Droplet data (as the Fake Night dataset has a lower FID than the Fake Droplet data (Section \ref{sec:fid})). This intuitively might be due to the fact that night images, by nature, are captured with lower reflective light than images taken during day-time, resulting in images that have lower number of colours and details than images captured during the day. This, in turn, makes it easier for generative models like CycleGAN to generate them because, again, they have fewer colours and details than day-time images.

 \vspace{-2mm}


\subsection{Detector behaviour using synthetic adverse conditions}

When models are trained on augmented (Fake) datasets, their performance difference on Real Sunny varies. the performance of YOLOv3 and YOLOV5 experienced marginal (negligible) improvement or degradation. However, ScaledYOLOv4 and EfficientDet performance was slightly degraded. 

The different detection models use a varied architecture for the neck structure (which is placed between the head and backbone to aggregate as much information extracted by the backbone as possible before it is fed to the head) - YOLOv3 and YOLOV5 use a Path Aggregation Network (PANet) \cite{liu2018path} as a neck and they exhibit a slight performance enhancement, while ScaledYOLOv4, FRCNN, and EfficientDet use FPN \cite{fpn2016} and they generally show a degradation of performance. The results suggest that detectors with an FPN neck are more susceptible to augmentations degrading good weather performance than those using PANet.

Among the five detectors, YOLOv5 is the most robust in all weather conditions - when trained on augmented data, it reported an improvement of 1.3, 52.1, and 7.6 pp on real sunny, real night, and real droplet respectively. On the other hand, EfficientDet performed the worst after augmentations, exhibiting a performance degradation of -8.4 and -5.5 pp for Sunny and Real Droplet and an improvement of +0.5 on Real Night. 

\subsection{Object Detectors' Performance}

Looking at Table \ref{tab:results}, it is observed that, in the absence of adverse weather training data (i.e., when only the sunny weather training dataset is available), Faster-RCNN is the most accurate model for sunny and rainy weather. However, YOLOv3-tiny is arguably the optimal model for this application, as object detection performance is only marginally reduced from that of Faster-RCNN in sunny and rainy conditions, but exhibits double the mAP in night-time tests - while simultaneously being 25 times faster and 18 times more computationally efficient than Faster-RCNN.

In contrast, when synthetic adverse condition training data is available, YOLOv5 and FRCNN appeared to be the best choices in terms of accuracy and robustness in all weather conditions. After training on synthetic adverse conditions (fake night and fake droplet), they scored the highest mAP across the board when tested on sunny weather, real night, and real droplet. YOLOv5 is also capable of achieving these results while running at a high frame rate of almost 14 fps. 

In terms of raw computational efficiency alone (last row of Table \ref{tab:results}), EfficientDet is the best model among the five detectors. It has the lowest number of floating point operations per second, which makes it ideal for low computational capacity hardware. Although EfficientDet has (5 times) lower FLOPS than YOLOv3, the latter is (2.8 times) faster than the former. This is because the term FLOPS describes the number of operations \textbf{per second} not the total number of operations. In other words, EfficientDet might have more operations in total than YOLOv3, but it divides and executes them over a longer time span (thus slower), ending up with a small number of operations per second. Having a lower FLOPs means the model is hardware-friendly and can be installed on relatively light processors including the popular NVIDIA Jetson. However, EfficientDet performance is decreased in almost all weather conditions with the use of augmented training data - suggesting it is a poor choice when using augmented training data in adverse conditions. 

Furthermore, Table \ref{tab:results} in terms of inference speed, YOLOv3 is the fastest model by a clear margin - triple the speed of EfficientDet, and an order of magnitude faster than SYOLOv4 and Faster-RCNN. Interestingly, the number of frames it can process per second is more than the summation of frames processed by all the other models combined. While running at this high speed, YOLOv3 achieved a similar performance to that of YOLOv5. Therefore, in case a lower powered processor hardware is to be used or a high frame rate is critical, YOLOv3 is the optimal choice. It is worth mentioning that the frame rates in Table \ref{tab:results} were measured while running the detectors on the AV racing PC mentioned in section \ref{sec:exp_set} which (at the time of writing) is 4 years old. State-of-the-art on-board hardware will further improve the performance of all detectors.


\section{Conclusions and future work}
\vspace{-0.4cm}

CycleGAN-based style transfer can be efficiently used to generate synthetic \textit{(fake)} adverse condition data that can in turn be used to improve the performance of object detection models for autonomous racing in \textit{real} adverse conditions. In this paper, training on synthetic data has improved the performance of four out of five state-of-the-art detection models by $42.7$ and $4.4$ percentage points when tested on \textit{real night} and \textit{real droplet} frames, respectively. Behaviour of detectors against augmentations varies; out of the five detectors, EfficientDet was the worst while YOLOv5 has emerged as arguably the optimal one as it managed to achieve high mAP of $97\%$, $74\%$ and $61\%$ for sunny, night and droplet respectively - while being very efficient, making it ideal for autonomous racing applications. It is noted that the models with PANet neck had positive response to augmentations than those with FPN neck. In applications where speed is of the essence, YOLOv3 is recommended as it stands as the fastest detection model. 

Results suggest that a significant improvement in object detection accuracy in adverse weather can be achieved by using a carefully selected combination of synthetic weather training data, and an appropriate object detector. It is believed that these findings will be applicable to on-road autonomous driving scenarios, however further work in this area is required.

An interesting area for further exploration is the influence of the various neck structures and feature extraction backbones employed in different object detectors, in order to truly understand and optimise the performance of object detection models for applications requiring the use of augmented training data.




\section{Acknowledgement}

The authors would like to thank the OBR Autonomous team for their input and assistance, and Tjeerd Olde Scheper, Peter Ball, Matthias Rolf and Gordana Collier for fruitful discussions and support throughout this work. This project has received funding from the European Union’s Horizon 2020 research and innovation programme, under grant agreement No. 964505 (E-pi). 


{\small
\bibliographystyle{ieee_fullname}
\bibliography{main}
}

\end{document}